# Three new sensitivity analysis methods for influence diagrams

**Debarun Bhattacharjya**
Business Analytics and Math Sciences
IBM T. J. Watson Research Center
Yorktown Heights, NY 10598, USA
debarunb@us.ibm.com

**Ross D. Shachter**
Management Science and Engineering
Stanford University
Stanford, CA 94305, USA
shachter@stanford.edu

## Abstract

Performing sensitivity analysis for influence diagrams using the decision circuit framework is particularly convenient, since the partial derivatives with respect to every parameter are readily available [Bhattacharjya and Shachter, 2007; 2008]. In this paper we present three non-linear sensitivity analysis methods that utilize this partial derivative information and therefore do not require re-evaluating the decision situation multiple times. Specifically, we show how to efficiently compare strategies in decision situations, perform sensitivity to risk aversion and compute the value of perfect hedging [Seyller, 2008].

## 1 INTRODUCTION

Decision making under uncertainty presents several opportunities and challenges beyond those encountered in pure reasoning under uncertainty. An influence diagram [Howard and Matheson, 1984] is a graphical model that represents the relationships between the decisions, uncertainties and preferences of a decision maker (DM), and it can be evaluated to reveal the optimal strategy and the DM's value for the decision situation. Decision circuits are graphical representations that not only enable efficient evaluation of influence diagrams [Bhattacharjya and Shachter, 2007; Shachter and Bhattacharjya, 2010], but also compute a wide array of sensitivity analysis results [Bhattacharjya and Shachter, 2008]. In this paper, we extend our previous work on sensitivity analysis and show how decision circuits can also be used to efficiently perform more sophisticated analysis such as strategy comparisons, sensitivity to risk aversion, and value of perfect hedging [Seyller, 2008] computations.

Sensitivity analysis refers to observing how the outputs of a system are affected as the inputs are varied. In decision situations, such analysis can provide the DM with insights, a better understanding of the situation and clarity of action. Decision circuits extend the research in arithmetic circuits [Darwiche 2000; 2003] towards evaluating and analyzing decision situations, and they reap similar benefits in efficiency.

For belief networks, efficient sensitivity analysis is feasible primarily due to the linear relationship between marginal probabilities and the parameters (see for instance Castillo et al [1997]; Kjaerulff and van der Gaag [2000]; van der Gaag and Renooij [2001]; Chan and Darwiche [2004]). On the other hand, there are unique challenges in sensitivity analysis for influence diagrams [Nielsen and Jensen, 2003; Bhattacharjya and Shachter, 2008] arising due to the non-linearities created by maximization operations for making decisions, as well as the non-linear utility function for any DM who is not risk neutral.

There are three main contributions in this paper. The first is a method for finding the value when the DM changes the policy for a particular decision while maintaining the other policies at the current optimal strategy. The DM may be interested in comparing the optimal strategy with the status quo or a strategy that is easier to implement. Second, we present an efficient way to perform sensitivity to risk aversion, which enables the DM to compare the optimal strategy with one that might be more flexible. Third, we show how to compute the value of perfect hedging for any uncertainty, which enables a risk-averse DM to value deals for reducing the "risk" in her current situation.

The common feature among our methods is that they all essentially deal with non-linear sensitivity analysis in some form. While they have previously been difficult to perform using other representations of decision situations, they can all be performed efficiently using partial derivatives that are easily obtainable in the decision circuit framework. In Section 2 we review the literature, while Sections 3, 4 and 5 explain the three methods respectively. Section 6 concludes the paper.

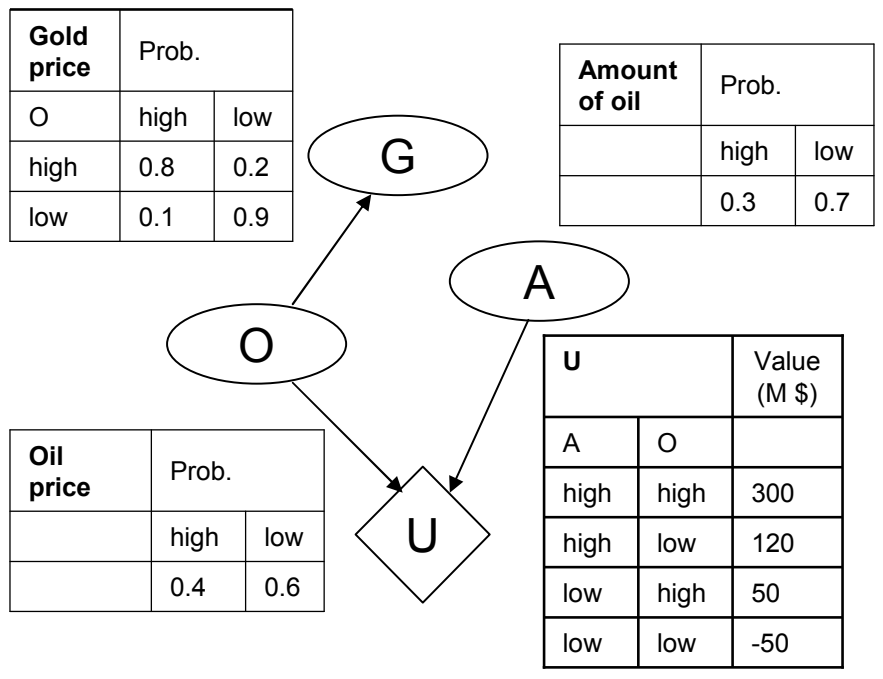

Figure 1: An influence diagram without decisions.

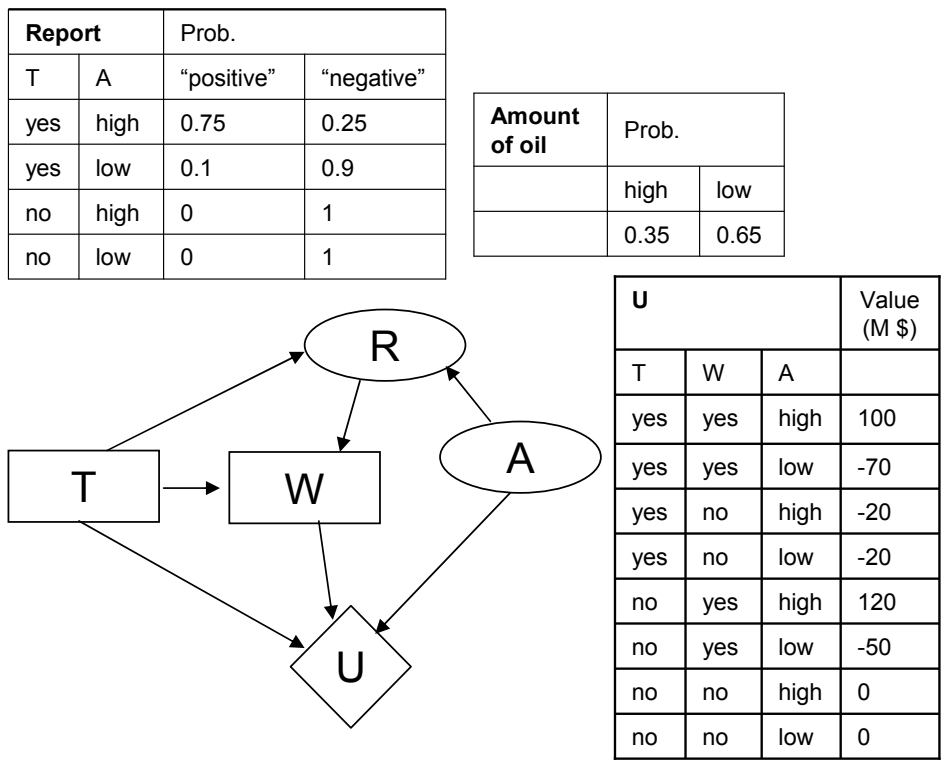

Figure 2: An influence diagram with decisions.

# 2 NOTATION AND REVIEW

Here we discuss notation, examples, and briefly review the key concepts. We assume that the reader is generally familiar with decision analysis fundamentals, including graphical models such as influence diagrams and belief networks.

## 2.1 Basic Notation and Examples

Variables are denoted by upper-case letters ($X$) and their values by lower-case letters ($x$). The *family* for $X$ also includes its parents $\mathbf{Pa(X)}$, and an instantiation of $X$'s family is denoted $x\mathbf{pa(X)}$. The bold-faced font indicates a set of variables. Whenever a variable $X$ is binary, we denote its states as $x$ and $\bar{x}$.

Consider the influence diagram in Figure 1 [Seyller, 2008]; in this case there are no explicit decisions in the model. The DM owns an oil field, and believes that the Amount of oil (A) is independent of the Oil price (O), but the Gold price (G) and Oil price are dependent. The utility node $U$ explicitly indicates that she will reap profits from oil production as determined by the Oil price and Amount of oil. We will discuss later why the DM might be concerned about the Gold price, even though it does not currently contribute to the value. In Figure 2 [Raiffa, 1968], the DM will decide whether to explore another field she owns with a seismic Test (T) to purchase a Report (R). Her testing decision and the report will be known to her when she decides whether to drill a Well (W). If she doesn't perform the test, the report is not informative and always states "negative".

In Figure 1, A and O are the *value attributes* since they are the parents $\mathbf{Pa(U)}$ of utility node $U$. We assess a *value table* $v(\mathbf{pa(U)})$ from a value function $v(.)$ that characterizes the value of the attributes in terms of a single numeraire, which we assume is dollars. We also assume that the analyst has assessed a utility function, $P(u|\mathbf{pa(U)}) = u(v(\mathbf{pa(U)}))$ where $u(.)$ is a von Neumann-Morgenstern utility function [von Neumann and Morgenstern, 1947] such that $u$ is at least as good and $\bar{u}$ is at least as bad as anything that can happen. Therefore we model $U$ as a binary node, where the utilities are normalized to lie in the interval $(0, 1)$.

The *certain equivalent* ($CE$) of an uncertain $V$, given by $u^{-1}(E[u(V)])$, represents the certain payment that the DM finds indifferent to $V$. Our sensitivity results will be expressed in terms of the $CE$ rather than utilities because the $CE$ is measured in the same units as the numeraire for value. Our chosen measure of value in this paper is dollars, but other numeraires are possible, such as quality adjusted life years (QALYs), number of lives saved, etc. Since utilities have an arbitrary scale and are used for the internal computations, they are meaningless to DM. We assume that the utility function $u(.)$ is strictly increasing and continuously differentiable. The most common utility functions are *linear*, $u(v) = u^0 v + u^\infty$, and *exponential*, $u(v) = -u^0 e^{-\gamma v} + u^\infty$, where $u^0 > 0$ and $\gamma > 0$ (we choose $u^0$ and $u^\infty$ such that $u(v) \in (0, 1)$). The exponential is the only utility function that captures risk-aversion and also allows us to express our valuation for a change in a decision situation as a difference between $CE$s with and without the change [Raiffa, 1968].

The conditional probability tables and the value function $v(.)$ are shown along with the influence diagrams in both Figures 1 and 2. We assume that the DM has an exponential utility function $u(.)$ with risk aversion $\gamma = 0.002$ (in units of $(M\$)^{-1}$) for both examples. Her $CE$s for Figures 1 and 2 are \$38.99$M$ and \$5.21$M$ respectively. In Figure 2, the optimal strategy is to test and drill the well when the report is "positive" and not to drill when the report is "negative".

## 2.2 Arithmetic and Decision Circuits

We can associate any belief network with a unique multi-linear function (MLF) over *evidence indicators*

$\lambda_x$, and *network parameters* $\theta_{x|\mathbf{pa}(\mathbf{X})}$. An arithmetic circuit [Darwiche, 2000; 2003] is a rooted, directed acyclic graph whose leaf nodes are either constants, $\lambda_x$ or $\theta_{x|\mathbf{pa}(\mathbf{X})}$, and all other nodes represent summation or multiplication. It can compactly factorize the MLF for efficient inference and sensitivity analysis.

An evidence indicator $\lambda_x$ is binary (0-1) with $\lambda_x = 0$ whenever $X$ has been observed taking another value, i.e. it is not $x$. There is an evidence indicator associated with each possible instantiation $x$ of each network variable $X$. A network parameter $\theta_{x|\mathbf{pa}(\mathbf{X})}$ represents a conditional probability, $\theta_{x|\mathbf{pa}(\mathbf{X})} = P(x|\mathbf{pa}(\mathbf{X}))$. There is a network parameter for each possible instantiation $x\mathbf{pa}(\mathbf{X})$ of family $X\mathbf{Pa}(\mathbf{X})$. Each term in the MLF corresponds to an instantiation $\mathbf{z}$ of all the network variables $\mathbf{Z}$, involving the product of all evidence indicators and network parameters consistent with $\mathbf{z}$. The MLF for a belief network is $f = \sum_{\mathbf{z}} \prod_{x\mathbf{pa}(\mathbf{X})\sim\mathbf{z}} \lambda_x \theta_{x|\mathbf{pa}(\mathbf{X})}$, where the sum is over every instantiation of all variables in the network and $x\mathbf{pa}(\mathbf{X}) \sim \mathbf{z}$ represents all families consistent with $\mathbf{z}$.

As an example, if we treat the utility node in Figure 1 as an uncertainty, the MLF is: $\sum_{a,o,g,u} \lambda_a \lambda_o \lambda_g \lambda_u \theta_a \theta_o \theta_{g|o} \theta_{u|a,o}$. An arithmetic circuit represents one particular factorization of this MLF, such as say: $\sum_a \lambda_a \theta_a \sum_o \lambda_o \theta_o \sum_g \lambda_g \theta_{g|o} \sum_u \lambda_u \theta_{u|a,o}$. There may be many possible factorizations (and therefore circuits) for a given MLF.

An arithmetic circuit performs inference by ensuring that the appropriate terms in the joint distribution are summed. By setting the evidence indicators to 0 or 1 such that they are consistent with the evidence $\mathbf{e}$, we can compute $P(\mathbf{e})$ in an *upward pass*, starting from the leaves and ending at the root. The result of *evaluating* the circuit in this upward pass is denoted as $f(\mathbf{e})$, where $f(\mathbf{e}) = P(\mathbf{e})$. We can calculate partial derivatives by *differentiating* the circuit through a subsequent *downward pass*, in which parents are visited before children. The upward and downward passes are also referred to as *sweeps*. The partial derivatives of the root with respect to every evidence indicator and network parameter are computed through the downward sweep. Darwiche [2003] shows how these derivatives can answer other inference queries such as conditionals, and also perform sensitivity analysis. We use the following result in particular for some proofs.

**Lemma 1.** *For every variable $X \notin \mathbf{E}$:*

$$P(x, \mathbf{e}) = \frac{\partial f}{\partial \lambda_x}(\mathbf{e})$$

For influence diagrams, Bhattacharjya and Shachter [2007] introduce decision circuits: arithmetic circuits augmented with maximization nodes. They represent the dynamic programming function corresponding to a sequential decision problem. The *size* of a decision circuit is the number of edges it contains.

For the decision nodes in an influence diagram, the evidence indicator $\lambda_d$ for decision $D$ is initialized to 0 only if the alternative $d$ is no longer available to the DM. We assume that each decision has at least two alternatives. The network parameter $\theta_{d|\mathbf{pa}(\mathbf{D})}$ for a decision $D$ is initialized to 1 if the alternative is conditionally available under scenario $\mathbf{pa}(\mathbf{D})$, and 0 otherwise. Any influence diagram can be transformed to a diagram with only one utility node $U$; the parameters $\theta_{u|\mathbf{pa}(\mathbf{U})} = u(v(\mathbf{pa}(\mathbf{U})))$ are the utilities (normalized to lie between 0 and 1), and $\theta_{\bar{u}|\mathbf{pa}(\mathbf{U})} = 1 - \theta_{u|\mathbf{pa}(\mathbf{U})}$.

An influence diagram represents a total ordering of the decisions, which enforces a partial order on the uncertainties. The order of maximization and summation in the dynamic programming function must respect this order. For the influence diagram in Figure 2, the only valid order is $T \prec R \prec W \prec A \prec U$, and one possible factorization is: $\max_t \lambda_t \theta_t \sum_r \max_w \lambda_w \theta_{w|tr} \sum_a \lambda_r \theta_{r|ta} \lambda_a \theta_a$ $\sum_u \lambda_u \theta_{u|trw}$.

Once compiled, a decision circuit can be evaluated in an upward sweep analogous to arithmetic circuits. Decision circuits are evaluated with evidence $\mathbf{e}' = \{\mathbf{e}, u\}$ when $\mathbf{e}$ is observed. The best outcome $u$ is also deemed to be observed in $\mathbf{e}'$ since the goal is to find optimal policies that maximize the probability of the best outcome given the evidence. The value of the root node of the circuit is denoted $g(\mathbf{e}')$. The optimal policies for all decisions are computed on the upward sweep at the maximization nodes, where the alternative $d^*$ with the highest value is chosen, breaking ties arbitrarily. The network parameter $\theta_{d|\mathbf{pa}(\mathbf{D})}$ is set to 0 for all other alternatives $d$. The optimal strategy $s*$ is the set of optimal policies $\theta_{d*|\mathbf{pa}(\mathbf{D})}$ for all decisions.

The circuit can be differentiated in a subsequent downward sweep, by treating max nodes as sum nodes. The decision circuit's root is the unnormalized expected utility; it can be normalized by the probability of evidence $P(\mathbf{e})$, computed by sweeping up the circuit with evidence $\mathbf{e}$ at the optimal strategy, $P(\mathbf{e}) = g(\mathbf{e}|s*)$, although it could be at any strategy since $\mathbf{e}$ is not responsive to the decisions. (Note that $P(\mathbf{e})$ can also be computed using derivatives from a downward sweep, $P(\mathbf{e}) = \frac{\partial g}{\partial \lambda_u}(\mathbf{e}'|s*) + \frac{\partial g}{\partial \lambda_{\bar{u}}}(\mathbf{e}'|s*))$. Thus the expected utility and $CE$ of the optimal strategy are:

**Lemma 2.** *The maximum EU is:*

$$EU(s*) = \frac{g(\mathbf{e}'|s*)}{g(\mathbf{e}|s*)}$$

**Lemma 3.** *For utility function $u(.)$, the $CE$ is:*

$$CE(s*) = u^{-1}\left(EU(s*)\right)$$

Once a decision circuit is compiled, it can be made even more compact by *pruning* [Bhattacharjya, 2008], i.e. exploiting deterministic relationships in the model to remove nodes and arcs that are not required from the circuit. For instance, in Figure 2, note that the Report always reads "negative" if the Test is not performed. Figure 3 shows a part of the pruned decision circuit for this influence diagram and the asymmetry that results from pruning is apparent.

### 2.3 Sensitivity Analysis in Influence Diagrams

Sensitivity analysis helps drive the conversation between the DM and the analyst. Howard [1968] uses terminology from the systems analysis literature to classify sensitivity analysis into two categories. He refers to varying a parameter and computing the certain equivalent at a fixed strategy as *open loop* analysis; when the strategy is allowed to vary by re-evaluating the decision situation, it is *closed loop* analysis. Closed loop results require re-evaluating the influence diagram, which can be done as efficiently as possible using decision circuits. Nonetheless, open loop results can provide the DM with key insights, using information that is already computed in the original circuit.

Previous work in sensitivity analysis in influence diagrams has involved varying parameters such as conditional probabilities and finding intervals over which the current strategy remains optimal [Nielsen and Jensen, 2003; Bhattacharjya and Shachter, 2008], drawing one-way sensitivity plots and computing the value of information [Howard, 1966; Raiffa, 1968]. In this paper we focus our attention on certain kinds of sensitivity analysis where non-linearity is a critical issue.

### 2.4 The Value of Perfect Hedging

The notion of hedging is well known particularly in the financial community and is associated with trading derivatives due to the long common history they share. Seyller [2008] writes: " ... in late 17th century Japan, a futures market in rice was developed at Dojima, near Osaka ... It was the first recorded instance in which such a market was created ". Seyller defines the value of hedging from a decision-analytic perspective, and it is this approach that we use in this paper.

In a decision situation, hedging occurs through a deal relevant to one we already possess. Suppose we own a coin toss deal where we receive \$50 for heads or pay \$50 for tails. Now consider another deal where the payoffs are reversed, i.e. we pay for heads and receive for tails. If we believe heads and tails are equally likely, then both these deals have identical value when considered separately. Moreover, being risk averse, we have a negative certain equivalent for either deal. However, if we were to own both deals, we would neither lose nor gain money for certain, regardless of the outcome of the coin toss! Thus, if we already owned one of these deals, we should be willing to pay for the other. Any risk averse DM can increase the value of his/her decision situation by purchasing other relevant deals.

The value of perfect hedging (VoPH) [Seyller, 2008] is a relatively new tool for the decision analyst, to help the DM choose deals that can help reduce her "risk" based on her current portfolio. The VoPH on uncertainty $X$ is defined as the most that the DM should be willing to pay for the best possible deal $Y$ that is determined solely by $X$ (i.e. $Y$ is a deterministic function of $X$), such that $E[Y] = 0$. For an important special case, the following holds true:

**Lemma 4.** *For a risk averse DM with an exponential utility function, a decision situation with evidence* $\mathbf{e}$ *and uncertainty* $X$,

$$VoPH(X|\mathbf{e}) = CE^{PH} - CE(s*),$$

*where* $CE^{PH}$ *is the CE with the perfect hedge, i.e. the maximum CE when any deal* $Y$ *(determined solely by* $X$*) is added to the current situation, under the constraint that* $E[Y|\mathbf{e}] = 0$*, and* $CE(s*)$ *is the CE of the original situation, without hedging.*

VoPH is defined to parallel the value of information (VoI). A DM should not be willing to pay more than VoI(X) for any information on uncertainty $X$. Similarly, a DM should not be willing to pay more than $E[Y] + VoPH(X)$ for any deal $Y$ determined solely by $X$. One major difference between VoI and VoPH is that while information is valuable only when it can affect the DM's choice, hedging can be valuable even without changing any decision since it affects the distribution of value. VoPH has several intuitive properties, e.g. it is non-negative for a risk-averse DM and zero for a risk-neutral DM or when the uncertainty is not relevant to the value. We will explore VoPH further in Section 5.

## 3 COMPARING STRATEGIES

Decision circuits contain indicators $\lambda_d$ and network parameters $\theta_{d|\mathbf{pa(D)}}$ for decisions. The power of the indicator $\lambda_d$ comes from being able to compute the value of an alternative (VoA), i.e. the minimum price that the DM should be willing to accept to forego the alternative in the decision situation, since setting $\lambda_d = 0$ removes alternative $d$ from consideration.

At first glance, it may appear that the derivative $\frac{\partial g(\mathbf{e}'|s*)}{\partial \lambda_d} = 0$ captures $VoA(d)$, but this is not true - the optimal policies have already been set to $s*$ and

the derivative is unable to compare the best alternative with the next best alternative. If $\frac{\partial g(\mathbf{e}'|s*)}{\partial \lambda_d} = 0$, there is 0 probability of this alternative being chosen at the current optimal strategy $s*$. If $\frac{\partial g(\mathbf{e}'|s*)}{\partial \lambda_d} = g(\mathbf{e}'|s*)$, this alternative is optimal for all observations $\mathbf{pa(D)}$ at $s*$, that is, alternative $d$ is always chosen in $s*$. For intermediate values, the derivative is an unnormalized product of utility and probability, and it may not be particularly meaningful to the DM. Instead, we suggest using the partial derivatives with respect to $\theta_{d|\mathbf{pa(D)}}$ for strategy comparisons.

Consider the following sensitivity analysis question: what is the $CE$ when the policy for only one decision is changed, while maintaining the policies for other decisions at the current optimal strategy $s*$? There are several reasons why the DM may be interested in such an analysis. For instance, the proposed optimal strategy may be different from the status quo, or a strategy that is easier to implement, and therefore the DM may wish to understand the incremental value from the optimal strategy. We show how this question can be answered through partial derivatives.

**Theorem 1.** *If we change the current optimal policy for decision $D$, denoted $d*(\mathbf{pa(D)})$, to $d'(\mathbf{pa(D)})$ keeping all other policies fixed to optimal strategy $s*$, then the $CE$ of the new strategy $s'$ is:*

$$CE(s') = u^{-1}\left(\frac{1}{g(\mathbf{e}|s*)}\left[\sum_{\mathbf{pa(D)}} \frac{\partial g(\mathbf{e}'|s*)}{\partial \theta_{d'|\mathbf{pa(D)}}}\right]\right)$$

*Proof.* During differentiation on the downward sweep, the max nodes are treated as sum nodes. For any strategy $s$, we can write the root value of the circuit as a linear function of the policies for decision $D$: $g(\mathbf{e}'|s) = \sum_d \sum_{\mathbf{pa(D)}} \theta_{d|\mathbf{pa(D)}} \frac{\partial g(\mathbf{e}'|s)}{\partial \theta_{d|\mathbf{pa(D)}}}$.
The strategies $s*$ and $s'$ only differ by the values of $\theta_{d|\mathbf{pa(D)}}$. The derivative $\frac{\partial g(\mathbf{e}'|s)}{\partial \theta_{d|\mathbf{pa(D)}}}$ does not depend on any of the values of $\theta_{d|\mathbf{pa(D)}}$, because each term in the linear function contains only one instantiation of family $D\mathbf{Pa(D)}$, therefore $\frac{\partial g(\mathbf{e}'|s')}{\partial \theta_{d|\mathbf{pa(D)}}} = \frac{\partial g(\mathbf{e}'|s*)}{\partial \theta_{d|\mathbf{pa(D)}}}$.
The policy $d'(\mathbf{pa(D)})$ sets $\theta_{d'|\mathbf{pa(D)}} = 1$ and the other $\theta$s to 0, thus:
$g(\mathbf{e}'|s') = \sum_{\mathbf{pa(D)}} \frac{\partial g(\mathbf{e}'|s')}{\partial \theta_{d'|\mathbf{pa(D)}}} = \sum_{\mathbf{pa(D)}} \frac{\partial g(\mathbf{e}'|s*)}{\partial \theta_{d'|\mathbf{pa(D)}}}$.
The required result follows from Lemmas 2 and 3. □

From the nature of the dynamic programming solution procedure for influence diagrams, optimal policies for all decisions are computed in reverse order, from the last decision to the first. In decision circuits, this is done by maximization nodes as optimal policies are identified during the upward sweep. Therefore, when the policy for a decision $D$ is modified from strategy $s*$, all the future decisions are guaranteed to remain optimal since that is how they were determined for $s*$. However, the policies for earlier decisions may no longer remain optimal after the policy changes for $D$.

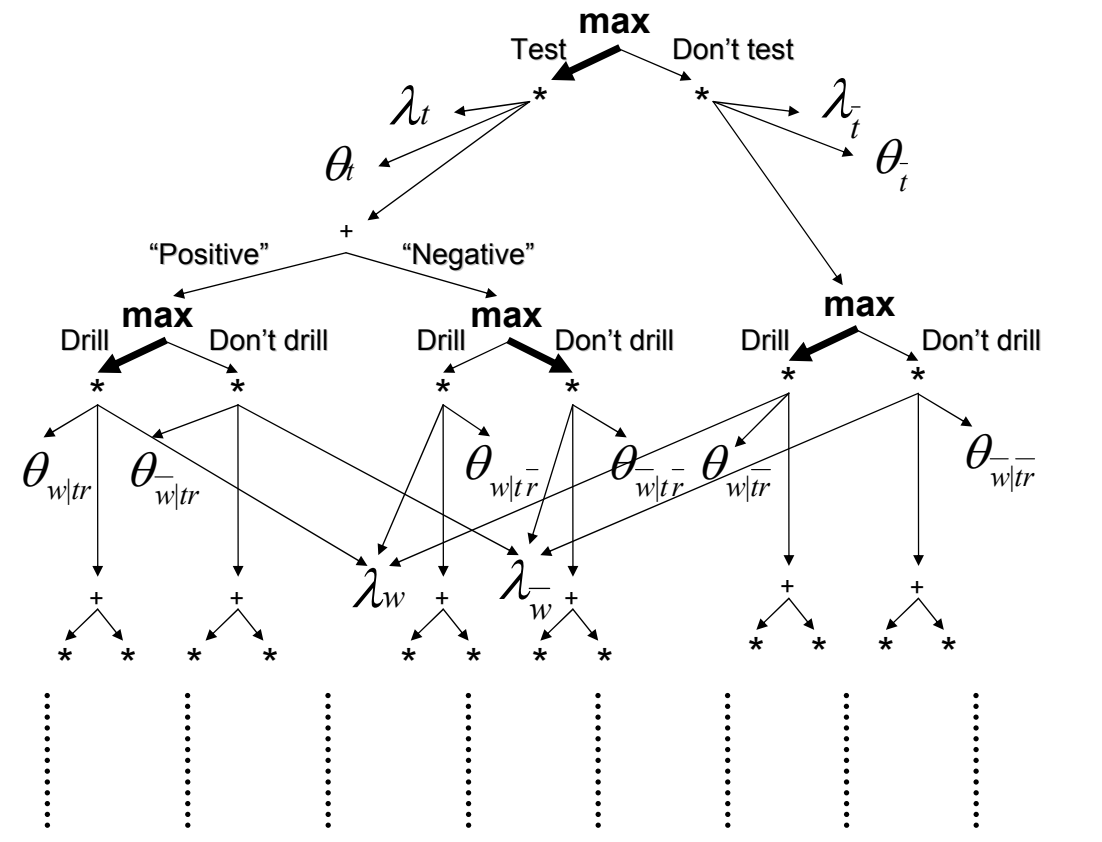

Figure 3: Partial decision circuit for the influence diagram shown in Figure 2.

Let us illustrate Theorem 1 with the partial decision circuit in Figure 3 (for the influence diagram shown in Figure 2). The current optimal strategy $s*$ is marked using bold arrows at the max nodes. There is no evidence in this case, therefore $\mathbf{e} = \emptyset$, and $\mathbf{e}' = \{\mathbf{e}, u\} = u$. $g(\mathbf{e}|s*) = g(\emptyset|s*) = 1$, $g(\mathbf{e}'|s*)$ equals the maximum expected utility (Lemma 2) and $CE(s*) = \$5.21M$ (Lemma 3).

Suppose the DM switches her choice at the top max node in Figure 3 to don't test, while maintaining the current optimal policy for the drilling decision. It is optimal to subsequently drill (see the bottom max node along the don't test path in Figure 3), which was determined by the initial dynamic programming solution procedure. For the new strategy (don't test, drill the well), $CE = u^{-1}\left(\frac{\partial g(\mathbf{e}'|s*)}{\partial \theta_{\bar{g}}}\right) = \$3.17M$. Now suppose the DM switches from $s*$ to always drilling regardless of what is observed. For the new strategy (test and drill regardless of what the report says), $CE = u^{-1}\left(\frac{\partial g(\mathbf{e}'|s*)}{\partial \theta_{w|tr}} + \frac{\partial g(\mathbf{e}'|s*)}{\partial \theta_{w|t\bar{r}}} + \frac{\partial g(\mathbf{e}'|s*)}{\partial \theta_{w|\bar{t}\bar{r}}}\right) = \$-16.83M$. This strategy is not optimal given the enforced drilling policy - it is clearly better not to perform the test, in which case the $CE$ would be $\$3.17M$ (as we noted earlier). The current optimal strategy is better than not testing and drilling by $\$(5.21 - 3.17)M = \$2.04M$. If there are some organizational concerns related to performing the test, such a strategy comparison can be beneficial for the DM.

Once the derivatives are available from initial sweeps through the decision circuit, finding the $CE$ by changing a policy using Theorem 1 is of the complexity of $O(|\mathbf{pa(D)}|)$ (where $|\mathbf{pa(D)}|$ is the number of possible observations for $D$). This is more efficient than sweeping through the circuit again, but it should be noted

that the earlier decisions are not re-optimized.

While earlier decisions may no longer remain optimal after a policy change, there are several potential applications of this sort of analysis: 1) DM may be at that decision epoch and therefore already made the earlier decisions; 2) There are practical conditions (like organizational reasons) which restrict the earlier decisions.

# 4 SENSITIVITY TO RISK AVERSION

It is often difficult to assess the DM's utility function, and therefore sensitivity analysis is an instructive technique for understanding the effects of risk attitude on the optimal strategy and the $CE$. The *local risk aversion* for utility function $u(.)$ is a measure of the DM's risk attitude at a specific dollar value, and is defined as $\gamma_v = -\frac{u''(v)}{u'(v)}$ [Pratt, 1964; Arrow, 1965]. The exponential utility function is the only function that features constant and non-zero risk aversion; in this case the local risk aversion is also the global risk aversion (i.e. it does not depend on $v$) and is denoted as $\gamma$. We will focus our attention on this important special case.

For small decision problems, $CE$ is typically plotted against $\gamma$ by re-evaluating the decision problem multiple times. Sensitivity to risk aversion for larger decision problems is challenging primarily due to two reasons: 1) A change in $\gamma$ changes all the entries $\theta_{u|\mathbf{pa}(\mathbf{U})}$; 2) It is no longer possible to make comparisons in the utility-space since the utility function itself is modified. However, there is a linear property when only the utilities are varied, similar to the case when only the parameters within the same conditional probability table are varied [Chan and Darwiche, 2004]:

**Lemma 5.** *The root of the decision circuit evaluated at evidence* $\mathbf{e}'$ *can be written as:*

$$g(\mathbf{e}'|s*) = \sum_{\mathbf{pa}(\mathbf{U})} \frac{\partial g(\mathbf{e}'|s*)}{\partial \theta_{u|\mathbf{pa}(\mathbf{U})}} \theta_{u|\mathbf{pa}(\mathbf{U})},$$

*where the partial derivatives* $\frac{\partial g(\mathbf{e}'|s*)}{\partial \theta_{u|\mathbf{pa}(\mathbf{U})}}$ *do not depend on the utilities.*

We can now compute the partial derivative of $CE(s*)$ with respect to any parameter $\nu$ of the utility function:

**Theorem 2.** *The partial derivative of* $CE(s*)$ *w.r.t any parameter* $\nu$ *of any utility function* $u(.)$ *is:*

$$\frac{\partial CE(s*)}{\partial \nu} = \left( \frac{\frac{\partial u^{-1}}{\partial \nu}|_{EU(s*)}}{g(\mathbf{e}|s*)} \right) \sum_{\mathbf{pa}(\mathbf{U})} \frac{\partial g(\mathbf{e}'|s*)}{\partial \theta_{u|\mathbf{pa}(\mathbf{U})}} \frac{\partial u}{\partial \nu}|_{v(\mathbf{pa}(\mathbf{U}))}$$

*Proof.* From Lemmas 2, 3 and 5,
$CE(s*) = u^{-1} \left( \frac{1}{g(\mathbf{e}|s*)} \sum_{\mathbf{pa}(\mathbf{U})} \frac{\partial g(\mathbf{e}'|s*)}{\partial \theta_{u|\mathbf{pa}(\mathbf{U})}} \theta_{u|\mathbf{pa}(\mathbf{U})} \right)$.
Parameter $\nu$ only affects utilities, not $g(\mathbf{e}|s*)$. The result follows from the chain rule of differentiation. □

The partial derivative from Theorem 2 yields sensitivity to risk aversion for the exponential utility function when $\nu = \gamma$, and it depends on readily available partial derivatives. Note that $CE(s*)$ is a non-linear function of $\gamma$, and while the derivative can be used for computing the local variation, it cannot entirely determine the plot across a significant range. However, we show next that there is a closed-form (non-linear) expression for the $CE(s*)$ as a function of $\gamma$, in terms of partial derivatives with respect to the utilities.

**Theorem 3.** *For a DM with an exponential utility function:*

$$CE(s*) = \frac{1}{\gamma} \left[ \ln \left( \left( \sum_{\mathbf{pa}(\mathbf{U})} \frac{\partial g(\mathbf{e}'|s*)}{\partial \theta_{u|\mathbf{pa}(\mathbf{U})}} e^{-\gamma v(\mathbf{paU})} \right) / g(\mathbf{e}|s*) \right) \right]$$

*Proof.* $\theta_{u|\mathbf{pa}(\mathbf{U})} = u^0 e^{-\gamma v(\mathbf{pa}(\mathbf{U}))} - u^\infty$.
From Lemma 3, $u(CE(s*)) = EU(s*)$.
In the expression above,
$LHS = u^0 e^{-\gamma CE(s*)} - u^\infty$.
From Lemmas 2 and 5,
$RHS = \frac{\left( \sum_{\mathbf{pa}(\mathbf{U})} \frac{\partial g(\mathbf{e}'|s*)}{\partial \theta_{u|\mathbf{pa}(\mathbf{U})}} \left( u^0 e^{-\gamma v(\mathbf{pa}(\mathbf{U}))} - u^\infty \right) \right)}{g(\mathbf{e}|s*)}$.
The result follows from recognizing that $\sum_{\mathbf{pa}(\mathbf{U})} \frac{\partial g(\mathbf{e}'|s*)}{\partial \theta_{u|\mathbf{pa}(\mathbf{U})}} = g(\mathbf{e}|s*)$, taking the natural log of both sides and rearranging terms. □

Theorem 3 is a powerful result that provides a closed-form expression capturing sensitivity to $\gamma$. Once the derivatives are available from initial sweeps through the decision circuit, plotting a graph of $CE(s*)$ vs. $\gamma$ is of the complexity of $O(R|\mathbf{pa}(\mathbf{U})|)$ where $R$ is the number of points over which the plot is made and $|\mathbf{pa}(\mathbf{U})|$ is the number of instantiations of the value attributes. This is much more efficient than re-evaluating the decision situation, which would require $O(R(dc))$ computations, where $dc$ is the size of the decision circuit. The caveat is that this is computed at the original optimal strategy, whereas re-evaluating the decision situation at a different $\gamma$ could yield a different optimal strategy.

While sensitivity to risk aversion is crucial in and of itself, Theorem 3 is particularly applicable for studying the *flexibility* of strategies. Shachter and Mandelbaum [1996] suggest that a flexible plan must be able to perform well under unanticipated or unmodeled uncertainty. Under several assumptions and for a DM with an exponential utility function, they prove that a more flexible solution is equivalent to a solution that is preferred for a larger risk aversion; therefore being capable of dealing with unmodeled uncertainties is equivalent to choosing a less "risky" solution.

Let us demonstrate this with an example, using the influence diagram from Figure 2, which was evaluated at $\gamma = 0.002$. The DM re-evaluates the influence diagram at a lower value, $\gamma = 0$ (risk-neutral), and at a higher value, $\gamma = 0.01$. At $\gamma = 0$ it is optimal to drill without testing; at $\gamma = 0.01$ it is optimal neither to test nor to drill. As she plots these three strategies with risk aversion (Figure 4) using Theorem 3, she can compare her current strategy with one that is more flexible and one that might be less flexible. Apart from the crossover points in the figure, she can also note that the difference between the $CE$s of the current optimal and the more flexible strategy at her currently assessed $\gamma = 0.002$ is: $\$(5.21-0)M = \$5.21M$. She should consider whether she is willing to pay that much for the gain in flexibility.

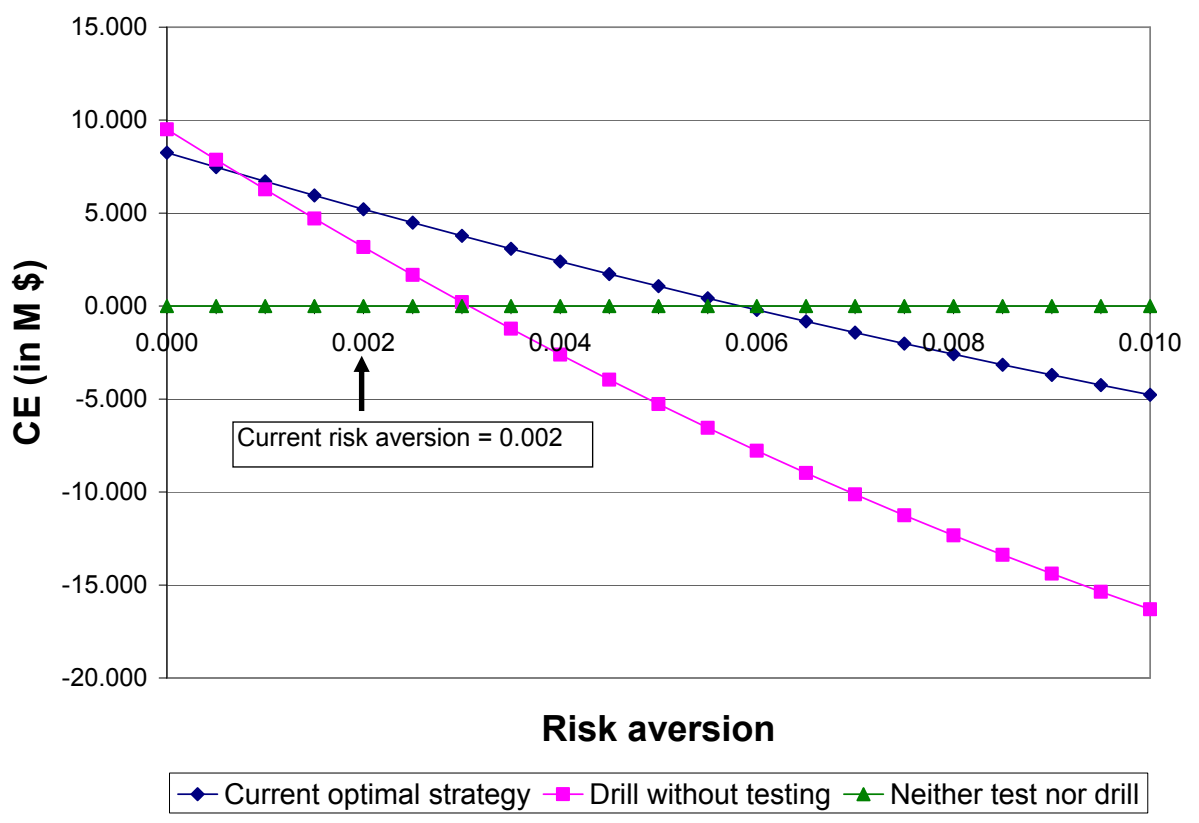

Figure 4: Sensitivity analysis with respect to risk aversion.

## 5 THE VALUE OF PERFECT HEDGING

The value of perfect hedging (VoPH), reviewed earlier in Section 2.4, is a compelling new technique for appraising the decision situation. It supports a proactive DM in evaluating the worth of other deals that are relevant to the current decision situation and can offset "risk" from the existing distribution of value.

Although the VoPH is a useful concept, computing it can be extremely challenging. From Lemma 4, note that finding the perfect hedge and the VoPH entails solving a difficult non-linear optimization problem, since $CE^{PH}$ is a non-linear function of the hedge. Seyller [2008] employs an exhaustive search method by sampling over several possible values of the perfect hedge. The theme in this paper is to avoid re-evaluating the decision situation for computations; through the theorems in this section, we show that there is a very efficient way to compute the VoPH at a fixed strategy using partial derivative information.

**Theorem 4.** *Consider a decision situation with evidence* $\mathbf{e}$ *and exponential utility function. The perfect hedge* $Y$ *(determined solely by* $X$*) at strategy* $s*$ *is:*

$$y(x) = E[CE(s*|X)] - CE(s*|x),$$

*and the VoPH is:*

$$VoPH(X|\mathbf{e}, s*) = E[CE(s*|X)] - CE(s*),$$

*where* $CE(s*|x)$ *is the conditional CE of strategy* $s*$ *given that* $x$ *is observed,* $E[CE(s*|X)]$ *is the probability weighted average of these conditional CEs, and* $CE(s*)$ *is the CE of the original situation.*

*Proof.* Let us partition the set of all uncertain variables into $X$ and $\mathbf{W}$. When we introduce additional value from $Y$, we change the utilities but the utility function remains the same, therefore maximizing certain equivalent is the same as maximizing expected utility. According to Lemma 4, the perfect hedge (when we observe evidence $\mathbf{e}$ and stay at the current optimal strategy $s*$) is the $y(x)$ that solves the following non-linear optimization problem:

$$\max_{y(x)} \sum_x P(x|\mathbf{e}, s*) \sum_{\mathbf{w}} P(\mathbf{w}|x, \mathbf{e}, s*) [u(v(x, \mathbf{w}) + y(x))]$$
$$\text{s.t.} \sum_x P(x|\mathbf{e}, s*) y(x) = 0$$

The objective function of the optimization problem is the expected utility at $s*$ with the perfect hedge, $EU^{PH}(s*)$. It is a concave function and the constraint is linear, therefore the necessary conditions are also sufficient. We first solve for the Lagrangian multiplier $\mu$, recognizing that $\sum_{\mathbf{w}} P(\mathbf{w}|x, \mathbf{e}, s*)u(v(x, \mathbf{w})) = EU(s*|x)$. Writing $y(x)$ as a function of $\mu$, we find:

$$y(x) = \frac{1}{\gamma} \ln(u^\infty - EU(s*|x)) - \frac{1}{\gamma} \ln\left(\frac{\mu}{\gamma}\right)$$

Replacing this in the constraint of the optimization problem, solving for $\mu$ and simplifying,

$$y(x) = \frac{1}{\gamma}\ln(u^\infty - EU(s*|x)) - \frac{1}{\gamma} \sum_x P(x|\mathbf{e}, s*) \ln(u^\infty - EU(s*|x))$$

Adding and subtracting the appropriate constant from both terms and from Lemma 3 we get the desired result. $VoPH(X|\mathbf{e}, s*)$ can be computed first by replacing $y(x)$ in the objective function to find $EU^{PH}$, and then using Lemma 3 to find $CE^{PH}$. The solution yields $CE^{PH} = E[CE(s*|X)]$. From Lemma 4, $VoPH(X|\mathbf{e}, s*) = E[CE(s*|X)] - CE(s*)$, which is the required solution. □

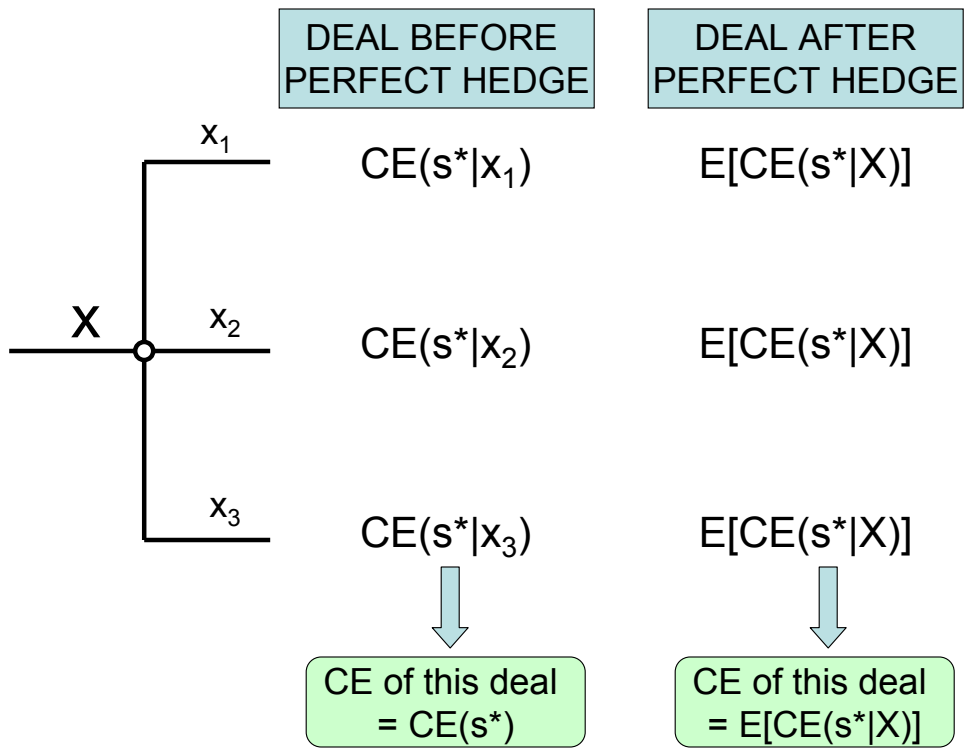

Figure 5: Before and after perfect hedging on $X$.

Let us explore the intuition behind the above result. Firstly, note that we have hidden the conditioning on the evidence $\mathbf{e}$ in the term $CE(s*|x)$ and it should be clear that all $CE$ computations are done conditioned on $\mathbf{e}$ (also for the original decision situation). Theorem 4 reveals that the perfect hedge $y(x)$ is the dollar value that adds or subtracts value to $CE(s*|x)$ to make the conditional $CE$ equal to $E[CE(s*|X)]$, for all states $x$. It is clear that $E[Y] = 0$ because $E[E[CE(s*|X)] - CE(s*|x)] = 0$. Perfect hedging on $X$ reveals an intuitive insight: the perfect hedge ensures receiving the expected value of the conditional $CE$s regardless of the outcome of $X$. Since the DM is risk averse, such a deal will never hurt the DM and thus $VoPH(X|\mathbf{e}, s*) \geq 0$. Figure 5 compares the deal before and after the perfect hedge, as we condition on some uncertainty $X$ with three states. $VoPH(X|\mathbf{e}, s*)$ is the difference between the $CE$s of the two situations.

The obvious next question is: how can we compute the perfect hedge and the VoPH efficiently? We propose using partial derivatives as follows.

**Theorem 5.** *Using terminology from the previous theorem,*

$$CE(s*|x) = u^{-1}\left(\frac{\partial g(\mathbf{e}'|s*)}{\partial \lambda_x} / \frac{\partial g(\mathbf{e}|s*)}{\partial \lambda_x}\right)$$

$$E[CE(s*|X)] = \frac{1}{g(\mathbf{e}|s*)}\left[\sum_x \frac{\partial g(\mathbf{e}|s*)}{\partial \lambda_x} CE(s*|x)\right]$$

*Proof.* The conditional expected utility $EU(s*|x) = P(u|x, \mathbf{e}, s*) = P(u, \mathbf{e}, x|s*)/P(\mathbf{e}, x|s*)$. From Lemma 1, the numerator is $\frac{\partial g(\mathbf{e}'|s*)}{\partial \lambda_x}$ and the denominator is $\frac{\partial g(\mathbf{e}|s*)}{\partial \lambda_x}$. The result for $CE(s*|x)$ follows from Lemma 3. From the definition of expected value, $E[CE(s*|X)] = \sum_x (P(x|\mathbf{e}, s*)CE(s*|x))$. Again from Lemma 1, $P(x|\mathbf{e}, s*) = \left(\frac{\partial g(\mathbf{e}|s*)}{\partial \lambda_x}\right)/g(\mathbf{e}|s*)$, which yields the required result for $E[CE(s*|X)]$. □

Table 1: VoPH results for the example in Figure 1.

| **Uncertainty** | **VoPH (in M \$)** |
|---|---|
| Amount of oil | 7.68 |
| Oil price | 3.26 |
| Gold price | 1.61 |

We see that the conditional $CE$s are very easy to compute using partial derivatives. Regarding the computational complexity of finding $VoPH(X|\mathbf{e}, s*)$, note that all the partial derivatives in Theorem 5 are available from four sweeps on the decision circuit: two sweeps, up and down with evidence $\mathbf{e}$ and another two sweeps, up and down with evidence $\mathbf{e}'$. Once we have obtained these derivatives, $VoPH(X|\mathbf{e}, s*)$ for any particular $X$ can be computed in constant time. Similar to the previous section, the caveat is that the solution is at a fixed strategy. In general, the current optimal strategy may no longer remain optimal after adding the perfect hedge, and therefore $VoPH(X|\mathbf{e}, s*) \leq VoPH(X|\mathbf{e})$.

The ideal application of Theorems 4 and 5 is when there are no decisions (as in Figure 1a) since we don't need to re-evaluate the decision situation. Suppose a DM (or organization) represents their complex portfolio by a belief network augmented with a utility node, to capture what is valued. In that situation, we can efficiently identify uncertainties that the DM should consider hedging. Moreover, an arithmetic circuit is sufficient (no maximization) and we can leverage the efficiencies of compact arithmetic circuits [Chavira, 2007].

Table 1 shows the VoPH for all three uncertainties in the influence diagram in Figure 1 [Seyller, 2008]. Suppose the DM is offered the following deal (perhaps a futures contract) on the Gold Price for $\$50M$: the payoff is $-\$50M$ when Gold price is high and $\$100M$ when Gold price is low. Then the VoPH informs the DM that this deal is not worth her perusal, because the price she should pay for this deal is bounded above by $E[Y] + VoPH(Y) = \$(43 + 1.61)M = \$44.61M$. Although this is a simple example, it is easy to see that the results scale up for large problems.

In general, we may wish to find the price that the DM is willing to pay for a specific deal that is determined by the value attributes in the model, not necessarily the perfect hedge. In the previous section, we used partial derivatives with respect to utilities to perform sensitivity to risk aversion since varying $\gamma$ changes all the utilities. A similar approach can be taken to find the $CE$ (at a fixed strategy) when this new deal is added to the original situation, because this addition only affects the utilities. Similar to Theorem 3, we can formulate a closed form expression for the new $CE$ as a function of the payoffs from this new deal and partial derivatives with respect to the utilities.

## 6 CONCLUSIONS

Sensitivity analysis is crucial in decision analysis, helping the DM consider the value of further information gathering and the benefits of obtaining greater parameter precision. In this paper, we presented three sensitivity analysis methods that can be easily applied in the decision circuit framework and do not require reevaluating the decision situation, making them highly accessible for potentially large decision problems. We have demonstrated how a DM should study the effect of switching to another policy or choosing a more flexible solution, as well as find deals that can effectively hedge his/her current portfolio. We have shown that DMs can gain valuable insights from the partial derivative information available in decision circuits.

### Acknowledgements

We thank Thomas Seyller, Léa Deleris and three anonymous reviewers for their valuable feedback.

### References

Arrow, K. J., 1965, The theory of risk aversion, In *Aspects of the Theory of Risk-Bearing*, Lecture 2, Yrjo Jahnssonin Saatio: Helsinki.

Bhattacharjya, D., and Shachter, R., 2007, Evaluating influence diagrams with decision circuits, In Parr, R., and van der Gaag, L., editors, *Proc. of 23rd UAI*, pp. 9-16, Vancouver, BC, Canada: AUAI Press.

Bhattacharjya, D., and Shachter, R., 2008, Sensitivity analysis in decision circuits, In McAllester, D., and Myllymaki, P., editors, *Proc. of 24th UAI*, pp. 34-42, Helsinki, Finland: AUAI Press.

Bhattacharjya, D., 2008, Decision circuits: A graphical representation for efficient decision analysis computation, Ph.D Thesis, Stanford University.

Castillo, E., Gutierrez, J. M., and Hadi, A. S., 1997, Sensitivity analysis in discrete Bayesian networks, *IEEE Transactions on Systems, Man, and Cybernetics*, **27**, pp. 412-423.

Chan, H., and Darwiche, A., 2004, Sensitivity analysis in Bayesian networks: From single to multiple parameters, In Chickering, M., and Halpern, J., editors, *Proc. of 20th UAI*, pp. 67-75, Banff, Canada: AUAI Press.

Chavira, M., 2007, Beyond treewidth in probabilistic inference, Ph.D Thesis, University of California - Los Angeles.

Darwiche, A., 2000, A differential approach to inference in Bayesian networks, In Boutilier, C., and Goldszmidt, M., editors, *Proc. of 16th UAI*, pp. 123-132, Stanford, CA, USA: Morgan Kaufmann.

Darwiche, A., 2003, A differential approach to inference in Bayesian networks, *Journal of the ACM*, **50** (3), pp. 280-305.

Howard, R., 1966, Information value theory, *IEEE Transactions on Systems Science and Cybernetics*, **2** (1), pp. 22-26.

Howard, R., 1968, The foundations of decision analysis, *IEEE Transactions on Systems Science and Cybernetics*, **4** (3), pp. 211-219.

Howard, R., and Matheson, J., 1984, Influence diagrams, In Howard, R., and Matheson, J., editors, *The Principles and Applications of Decision Analysis*, Vol. II, Strategic Decisions Group, Menlo Park, CA.

Kjaerulff, U., and van der Gaag, L. C., 2000, Making sensitivity analysis computationally efficient, In Boutilier, C., and Goldszmidt, M., editors, *Proc. of 16th UAI*, pp. 317-325, Stanford, CA, USA: Morgan Kaufmann.

Nielsen, T., and Jensen, F. V., 2003, Sensitivity analysis in influence diagrams, *IEEE Transactions on Systems, Man, and Cybernetics - Part A: Systems and Humans*, **33** (1), March, pp. 223-234.

Pratt, J., 1964, Risk aversion in the small and in the large, *Econometrica*, **32**, pp. 122-136.

Raiffa, H., 1968, *Decision Analysis: Introductory Lectures on Choices under Uncertainty*, Addison-Wesley.

Seyller, T., 2008, The value of hedging, Ph.D Thesis, Stanford University.

Shachter, R., and Bhattacharjya, D., 2010, Dynamic programming in influence diagrams with decision circuits, In *Proc. of 25th UAI*, Catalina Island, CA, USA.

Shachter, R., and Mandelbaum, M., 1996, A measure of decision flexibility, In Horvitz, E., and Jensen, F. V., editors, *Proc. of 12th UAI*, pp. 485-491, Portland, OR, USA: Morgan Kaufmann.

van der Gaag, L. C., and Renooij, S., 2001, Analysing sensitivity data from probabilistic networks, In Breese, J., and Koller, D., editors, *Proc. of 17th UAI*, pp. 530-537, Seattle, WA, USA: Morgan Kaufmann.

von Neumann, J., and Morgenstern, O., 1947, *Theory of Games and Economic Behavior*, 2nd edition, Princeton University Press, Princeton, NJ.